\documentclass{Interspeech2024}

\usepackage{verbatim}
\usepackage{booktabs}
\usepackage{hyperref}
\usepackage{multirow}
\usepackage{multicol}
\usepackage{comment}
\usepackage{subfig}
\usepackage{amsmath}
\usepackage{amssymb}
\usepackage[vlined]{algorithm2e}
\usepackage{url}
\usepackage{xcolor}
\usepackage{float}
\usepackage{algpseudocode}
\usepackage{graphicx}
\usepackage{booktabs}
\usepackage{subfig}

\usepackage{subcaption}
\usepackage{soul}
\usepackage[table]{colortbl}

\usepackage{multirow}
\usepackage{url}
\interspeechcameraready

\begin{document}
\title{Towards Multilingual Audio-Visual Question Answering}

\name[affiliation={1}]{Orchid}{Chetia Phukan*}
\name[affiliation={2}]{Priyabrata}{Mallick*}
\name[affiliation={2}]{Swarup Ranjan}{Behera*}
\name[affiliation={2}]{Aalekhya}{Satya Narayani}
\name[affiliation={1}]{Arun Balaji}{Buduru}
\name[affiliation={1,3}]{Rajesh}{Sharma}

\address{
  $^1$IIIT-Delhi, India,
  $^2$Reliance Jio AICoE, Hyderabad, India\\
  $^3$University of Tartu, Estonia\\
  *equal contribution}
\email{orchidp@iiitd.ac.in, priyabrata.mallick@ril.com, swarup.behera@ril.com}


\newcommand{\red}[1]{\textcolor{red}{#1}}

\maketitle

\begin{abstract}
In this paper, we work towards extending Audio-Visual Question Answering (AVQA) to multilingual settings. Existing AVQA research has predominantly revolved around English and replicating it for addressing AVQA in other languages requires a substantial allocation of resources. 
As a scalable solution, we leverage machine translation and present two multilingual AVQA datasets for 
eight languages created from existing benchmark AVQA datasets. This prevents extra human annotation efforts of collecting questions and answers manually. 
To this end, we propose, \textbf{MERA} framework, by leveraging state-of-the-art (SOTA) video, audio, and textual foundation models for AVQA in multiple languages. We introduce a suite of models namely MERA-L, MERA-C, MERA-T with varied model architectures to benchmark the proposed datasets. 
We believe our work will open new research directions and act as a reference benchmark for future works in multilingual AVQA.  
\end{abstract}

\vspace{-0.2cm}
\section{Introduction}

Humans naturally integrate visual and auditory stimuli, allowing for a holistic understanding of their environment. Audio-Visual Question Answering (AVQA) aims to emulate innate multimodal cognitive system of humans in machines. In a formal context, when provided with a video stream, AVQA endeavors to address natural language queries by fusing details from both auditory and visual modalities. Picture a video showing a calm beach scene, and a AVQA system is asked, ``What caused the sudden stir?'' To answer well, the system needs to blend what it sees - maybe waves crashing or people reacting - with what it hears, like the roar of an approaching storm or the cries of seagulls flying off. This shows how important it is for AVQA to understand how audio and visuals connect for accurate responses. This field of investigation has become increasingly prominent in recent years~\cite{yang2022avqa,li2022learning,yun2021pano, li2023progressive}.  However, developments in AVQA primarily concentrated on English, thus restricting its accessibility to a more advantaged segment of the global community.

One could contend that the prevalence of English in the field is largely attributed to the existence of AVQA benchmarks designed specifically in English \cite{yang2022avqa, li2022learning}. Creating these benchmarks takes a lot of resources, mainly because human annotators need to carefully collect and check the questions and answers for each videos. In this paper, we address the goal of extending AVQA to languages other than English by introducing two machine-translated multilingual AVQA datasets thus preventing extra human annotation effort. These datasets are generated by creating question-answer pairs in multiple languages based on two existing benchmark AVQA datasets: MUSIC-AVQA~\cite{li2022learning} and AVQA~\cite{yang2022avqa}. Our work follows a recent lineage of works that uses translation for data generation for multilingual Question-Answering (QA) systems \cite{changpinyo-etal-2023-maxm, behera23_interspeech}. Furthermore, we propose, \textbf{MERA} (\textbf{M}ultilingual Audio-Visual Question Answ\textbf{ER}ing Fr\textbf{A}mework) and it makes use of state-of-the-art (SOTA) video (VideoMAE \cite{tong2022videomae}), audio (AST \cite{gong21b_interspeech}), and textual (multilingual BERT \cite{devlin2018bert}) foundation models for effective multilingual AVQA. In summary, the main contributions are as follows.

    
    \begin{itemize}
        \item We present two multilingual AVQA datasets - m-MUSIC-AVQA and m-AVQA - in eight diverse languages: English (en), French (fr), Hindi (hi), German (de), Spanish (es), Italian (it), Dutch (nl), and Portuguese (pt). 
        \item We propose \textbf{MERA} framework for multilingual AVQA and suite of models namely, MERA-C, MERA-L, and MERA-T. MERA-C, MERA-L, MERA-T uses CNN, LSTM, and transformer networks respectively.  
        \item Extensive experiments with the proposed models show the usefulness of the newly created multilingual AVQA datasets and the effectiveness of the models towards providing correct answers. MERA-C showed the best performance in most instances for different languages and question types. 
        \item Furthermore, we show that weighted-ensemble of the models output probabilities leads to improvements across all the languages and different question types compared to the individual models. 
        
    \end{itemize}

    \noindent The dataset and code can be accessed at \footnote{\url{https://github.com/swarupbehera/mAVQA}}.

\vspace{-0.3cm}
\section{Related Work} \label{sec:relatedwork}
Here, we present a brief overview of past research encompassing various QA types and their efforts towards extending it to multilingual settings. 
We only focus on the multimodal QA (Visual Question Answering (VQA), Audio Question Answering (AQA), and AVQA) types that requires grounding natural language queries based on other modalities.  
\vspace{-0.3cm}
\subsection{Visual Question Answering}
\vspace{-0.2cm}
VQA generally involves training a system to comprehend the content of an image and respond to questions about it using natural language. Initially, Antol et al. \cite{antol2015vqa} presented a dataset comprising 204721 images sourced from the MS COCO dataset and an additional abstract scene dataset containing 50000 scenes. This dataset includes 760,000 questions and approximately 10 million corresponding answers. Furthermore, researchers proposed REVIVE \cite{lin2022revive} for effective VQA which uses a region-based approach on the input image that performs better in comparison to whole image-based and sliding window-based approaches. 
Changpingyo et al. \cite{changpinyo-etal-2023-maxm} made the first leap towards extending VQA for multiple languages and introducing MaXM, a multilingual benchmark for VQA in 7 languages.

\begin{table}[!t]
  \caption{Comparative overview of AVQA datasets. Sound type (ST) - O represents Object sounds. Visual Scene Type (VST), Number of videos (\#Video). Number of question-answer pairs (\#QA pairs).}
      \label{tab:AVQAdata}
  \centering
    \begin{tabular}{lcccc}
    \toprule
     \textbf{Dataset}       & \textbf{ST} & \textbf{VST} & \textbf{\#Video} & \textbf{\#QA pairs} \\
     \midrule
    MUSIC-AVQA & O & Music & 9.3k & 45.9K \\
    AVQA & O & Real-life & 57.0K & 57.3K \\
    Pano-AVQA & O & Panoramic & 5.4K & 51.7K \\
     \bottomrule
    \end{tabular}
\end{table}
\vspace{-0.3cm}
\subsection{Audio Question Answering}
\vspace{-0.2cm}
AQA is a task where a system interprets audio signals and natural language questions to produce a desired natural language output. AQA has seen substantial growth due to the availability of high-quality datasets, for example, ClothoAQA \cite{lipping2022clotho} which consisted of 1991 audio files with `yes' or `no' and single-word question-answer types. They proposed LSTM-based models for benchmarking clothoAQA. Fayek et al. \cite{fayek2020temporal} introduced the DAQA dataset for AQA consisting natural sound events. They then presented a novel framework called Multiple Auxiliary Controllers for Linear Modulation (MALiMo), which builds upon the Feature-wise Linear Modulation (FiLM) model. 
However, these studies were for English AQA and Behera et al. \cite{behera23_interspeech} led the first step towards multilingual AQA by proposing AQA dataset for eight languages and a BiLSTM-based framework. 

\begin{table}[!t]
   \caption{Analysis of question types and top five frequent answer candidates across MUSIC-AVQA and AVQA Datasets. Question types (QT). Total number of questions (\#Q). Number of questions for top 5 answers (\#Q5). 
   Top 5 classes of answers.}
  \label{tab:question_types}
  \centering
   \begin{tabular}{l|c|c|p{3.24cm}}
     \toprule
      \textbf{QT} & \textbf{\#Q} & \textbf{\#Q5} & \textbf{Top 5 Classes} \\
    \midrule
     \multicolumn{4}{c}{\textbf{MUSIC-AVQA}} \\
     \midrule
     Existential & 4990 & 4990 &  no, yes\\ \hline
    Location & 4615 & 2539 & yes, no, left, right, middle\\\hline
     Counting & 6351 & 6225 &  zero, one, two, three, four  \\\hline
     Comparative & 5545 & 5545 &  no, yes\\\hline
     \multirow{2}{*}{Temporal} & \multirow{2}{*}{4131} & \multirow{2}{*}{2080} & simultaneously, right, left, middle, violin\\
     \midrule
     \multicolumn{4}{c}{\textbf{AVQA}} \\
     \midrule
     \multirow{2}{*}{Which} & \multirow{2}{*}{22005} & \multirow{2}{*}{6707} & dog, bird, cat, chicken, cattle \\ \hline
     \multirow{3}{*}{Come from} & \multirow{3}{*}{16282} & \multirow{3}{*}{4217} & train, aircraft, sound of wind, motorcycle, helicopter \\ \hline
     \multirow{3}{*}{Happening} & \multirow{3}{*}{11346} & \multirow{3}{*}{1310} & rope skipping, skiing, rowing, riding, machine gun fire \\ \hline
     \multirow{2}{*}{Where} & \multirow{2}{*}{6805} & \multirow{2}{*}{1539} & at sea, field, eabed, highway, aquatic \\ \hline
     \multirow{3}{*}{Why} & \multirow{3}{*}{256} & \multirow{3}{*}{198} & i’m hungry, decompression, roller coaster ride, frightened, motorcycle \\ \hline
     \multirow{4}{*}{Before next} & \multirow{4}{*}{150} & \multirow{4}{*}{4} & volcanic explosion, setting off fireworks, tornado, sharpen the knife, set off firecrackers \\ \hline
     \multirow{2}{*}{When} & \multirow{2}{*}{64} & \multirow{2}{*}{21} & evening, chicken, train, lion, sound of wind \\ \hline
     \multirow{3}{*}{Used for} & \multirow{3}{*}{57} & \multirow{3}{*}{12} & decompression, train, dog, turkey, protect your eyes \\
     \bottomrule
     \end{tabular}
\end{table}

    \begin{figure*}[!tbh]
     \centering
       \includegraphics[width=\linewidth]{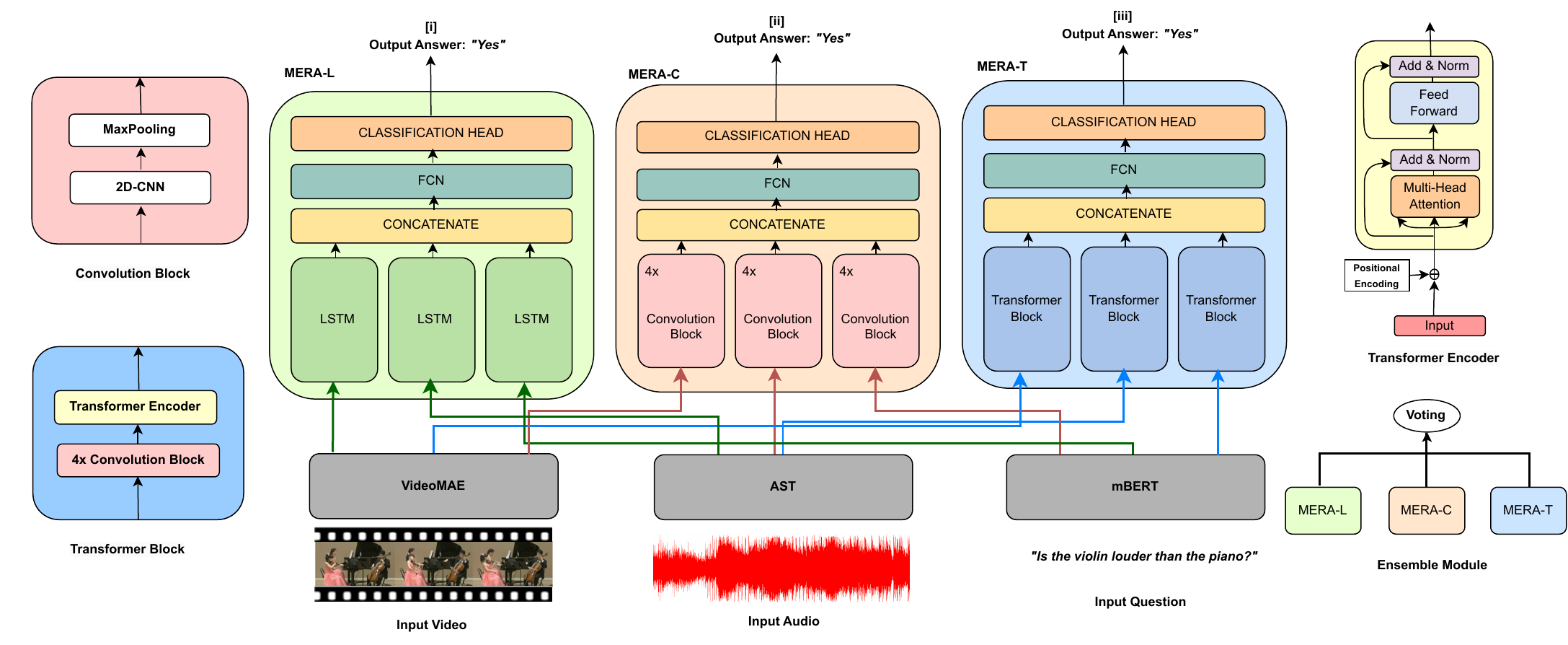}
      \caption{Proposed Framework, \textbf{MERA}; Here, \textbf{MERA} takes the video, audio, text (Question) as the input and answer is the output; The foundation models VideoMAE, AST, mBERT are kept frozen; Suite of three models namely MERA-L (i), MERA-C (ii), MERA-T (iii); Ensemble Module represents the weighted-ensemble of the three models}
      \label{fig:archi}
    \end{figure*}
\vspace{-0.3cm}    
\subsection{Audio-Visual Question Answering} 
\vspace{-0.20cm}
Yun et al. \cite{yun2021pano} proposed an innovative method for answering spatial and audio-visual questions in 360° videos, aiming for a comprehensive semantic understanding of omnidirectional surroundings. Yang et al. \cite{yang2022avqa} introduced AVQA, a real-life audio-visual dataset, and proposed a Hierarchical Audio-Visual Fusing (HAVF) model for the same. Furthermore, Li et al. \cite{li2022learning} proposed MUSIC-AVQA dataset and developed a novel spatiotemporal grounding model to tackle complex comprehension and reasoning tasks involving both audio and visual modalities. Researchers also propose an object-aware  approach that explicitly forms relation between the objects, sounds, and questions for improved AVQA on MUSIC-AVQA \cite{li2023object}. However, all these studies work only in English and AVQA in multilingual settings hasn't been explored yet. In this study, we work towards this direction.

    
\section{Multilingual AVQA Datasets}     \label{sec:dataset}

Here, in this section, we discuss the datasets curated in our study for multilingual AVQA. Acquiring high-quality labeled data remains a key challenge in multilingual AVQA, as in many other machine learning tasks. To address this, we introduce m-MUSIC-AVQA and m-AVQA datasets designed for multilingual AVQA. It is noteworthy to mention the existence of three datasets for AVQA: MUSIC-AVQA~\cite{li2022learning}, AVQA~\cite{yang2022avqa}, and Pano-AVQA~\cite{yun2021pano}. We present the details of the datasets, including sound type, visual scene type, number of videos, and number of question-answer pairs in Table~\ref{tab:AVQAdata}. It's crucial to emphasize that Pano-AVQA isn't publicly accessible. Therefore, we focus on the other two datasets for this study. For comprehensive statistical illustrations of the MUSIC-AVQA and AVQA datasets, including the distribution of video categories and the distributions of question-answer pairs, please refer to the respective papers. In Table~\ref{tab:question_types}, we provide an analysis of the question types alongside the top five frequent candidates for each type across both datasets.

\begin{figure}[!t]
      \centering
      \includegraphics[width=\linewidth]{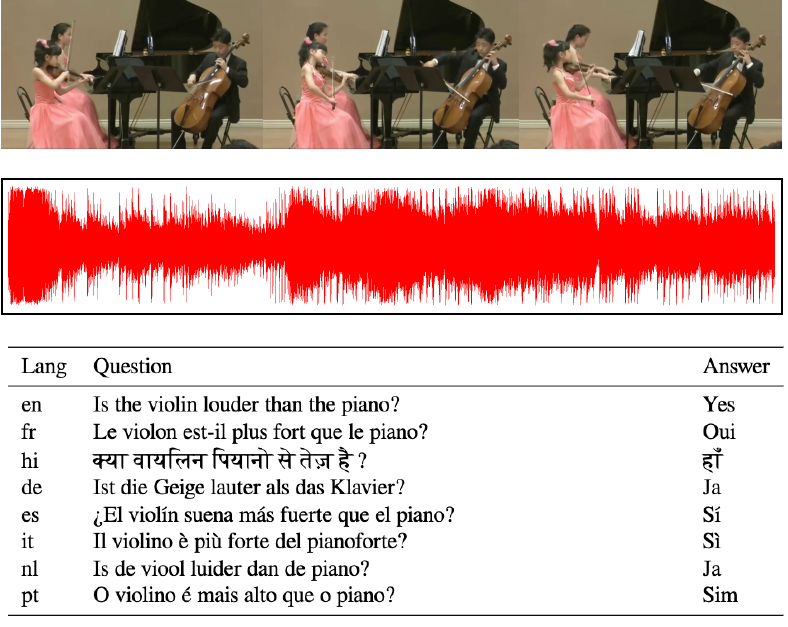}
      \caption{Multilingual MUSIC-AVQA (m-MUSIC-AVQA) dataset in eight languages. From top to bottom: English (en), French (fr), Hindi (hi), German (de), Spanish (es), Italian (it), Dutch (nl), and Portuguese (pt)}
      \label{fig:AVQAData}
    \end{figure}

To create multilingual datasets from MUSIC-AVQA and AVQA, we translated questions and answers from these datasets into seven additional languages using Google's machine translation API~\footnote{\url{https://cloud.google.com/translate}}. We also experimented with several other open-source machine translation tools, but their translation accuracy fell short of Google's. Evaluation with standard metrics such as BLEU \cite{papineni2002bleu}, ROUGE \cite{lin2004rouge}, and METEOR \cite{banerjeemeteor} confirmed the reliability of our translations. The selected languages include French (fr), Hindi (hi), German (de), Spanish (es), Italian (it), Dutch (nl), and Portuguese (pt). Furthermore, human verification and refinement were conducted on the translated question-answer pairs to ensure accuracy. We followed the multilingual AQA work by Behera et al. \cite{behera23_interspeech} for the selection of the languages for our study and in addition, they also reported that translation through Google's machine translation gave the best results. Examples of m-MUSIC-AVQA entries are provided in Figure~\ref{fig:AVQAData}. 



    

\vspace{-0.7cm}

\section{Methodology}
Here, we discuss various components of the proposed framework, \textbf{MERA} shown in Figure \ref{fig:archi}. First, we discuss the foundation models for extracting meaningful representations from different modalities followed by the suite of models proposed. We leverage the foundation models based on the SOTA performance in respective modalities for various tasks. We use VideoMAE, Audio Spectrogram Transformer (AST), multilingual BERT (mBERT) for extracting video, audio, and text representations respectively. 
\vspace{-0.3cm}
\subsection{Foundation Models}

\noindent\textbf{VideoMAE \cite{tong2022videomae}}: It is a revolutionary representation learning model for video trained in a self-supervised manner inspired by Image MAE \cite{he2022masked} that involves masked video pre-training. It is built upon simple ViT \cite{dosovitskiy2020image} backbones. VideoMAE showed superior performance in comparison to constrastive self-supervised pre-trained models in various downstream video tasks. 

\noindent\textbf{AST \cite{gong21b_interspeech}}: It is the first convolution-free fully attention-based model for audio classification. It uses pre-trained ViT \cite{dosovitskiy2020image} as the backbone architecture and further fine-tuned on AudioSet. 
AST showed SOTA performance for various audio classification tasks such as on ESC-50, speech commands, and so on.

\begin{table*}[htbp]
\scriptsize
\centering
\caption{Evaluation Scores of Models on different languages; 
L, C, T, ENS stands for MERA-L, MERA-C, MERA-T, and Ensemble respectively; A stands average; Values in \textcolor{blue}{\textbf{blue}} represents the highest score in a question type for a particular language and  \textcolor{blue}{\textbf{blue}} in A represents the highest score after averaging across all the languages for a question type; The scores are accuracy scores as used by \cite{li2022learning} and are in \%}
\begin{tabular}{p{0.15cm}|p{0.4cm} p{0.4cm} p{0.4cm} p{0.4cm} |p{0.4cm}p{0.4cm}p{0.4cm}p{0.4cm}|p{0.4cm}p{0.4cm}p{0.4cm}p{0.4cm}|p{0.4cm}p{0.4cm}p{0.4cm}p{0.4cm}|p{0.4cm}p{0.4cm}p{0.4cm}p{0.4cm}}

\toprule
\textbf{L} & \multicolumn{4}{c|}{\textbf{Existential}} & \multicolumn{4}{c|}{\textbf{Comparative}} & \multicolumn{4}{c|}{\textbf{Counting}} & \multicolumn{4}{c|}{\textbf{Location}} & \multicolumn{4}{c}{\textbf{Temporal}}\\ 
  & \textbf{L} & \textbf{C} & \textbf{T} & \textbf{ENS} & \textbf{L} & \textbf{C} & \textbf{T} & \textbf{ENS} & \textbf{L} & \textbf{C} & \textbf{T} & \textbf{ENS} & \textbf{L} & \textbf{C} & \textbf{T} & \textbf{ENS} & \textbf{L} & \textbf{C} & \textbf{T} & \textbf{ENS}  \\ \midrule
 \textbf{en} 
 & 79.09 & \textcolor{blue}{\textbf{81.50}} & 80.60 &  80.70
 & 54.08 & 60.00 & \textcolor{blue}{\textbf{61.43}} & 60.35
 & 55.72 & \textcolor{blue}{\textbf{59.18}} & 51.64 & 57.29
 & 35.16 & \textcolor{blue}{\textbf{47.57}} & 33.54 & 46.92
 & 35.43 & \textcolor{blue}{\textbf{51.69}} & 32.40  & \textcolor{blue}{\textbf{51.69}} \\ \midrule
 \textbf{fr} 
 & 80.80 & 80.60  & 80.20 & \textcolor{blue}{\textbf{81.50}} 
 & 59.46 & 60.26 & 59.19 & \textcolor{blue}{\textbf{61.25}}
 & 58.47 & 57.84  & 53.45 & \textcolor{blue}{\textbf{58.94}} 
 & 38.29  & \textcolor{blue}{\textbf{49.40}} & 33.44 & 48.43 
 & 43.20 & 52.66 & 31.67 & \textcolor{blue}{\textbf{54.36}} \\ \midrule
 \textbf{hi} 
 & 81.20 & 80.10 & 80.80 & \textcolor{blue}{\textbf{81.40}}
 & 57.93 & 62.06 & 60.89 & \textcolor{blue}{\textbf{63.31}} 
 & 59.18 & \textcolor{blue}{\textbf{60.51}} & 56.75 & 59.34 
 & 41.53 & \textcolor{blue}{\textbf{50.26}} & 34.84 & 49.94 
 & 43.68 & 51.82 & 32.28 & \textcolor{blue}{\textbf{53.15}}\\ \midrule
 \textbf{de} 
 & 79.89 & 77.38 & \textcolor{blue}{\textbf{80.60}} & 80.10
 & 58.20 & 62.24 & 59.91 & \textcolor{blue}{\textbf{63.58}} 
 & 59.65 & \textcolor{blue}{\textbf{60.28}} & 56.20 & 60.04 
 & 43.14 & 49.94 & 32.68 & \textcolor{blue}{\textbf{50.70}} 
 & 41.38 & 52.42 & 32.28 & \textcolor{blue}{\textbf{55.21}} \\ \midrule
 \textbf{es} 
 & 78.79 & 81.10 & 80.60 & \textcolor{blue}{\textbf{81.30}} 
 & 58.29 & 60.98 & 59.01 & \textcolor{blue}{\textbf{61.34}} 
 & 58.79 & \textcolor{blue}{\textbf{61.38}} & 55.72 & 61.30 
 & 36.89 & \textcolor{blue}{\textbf{45.73}} & 32.03 & \textcolor{blue}{\textbf{45.73}} 
 & 45.14 & 50.24 & 31.67 & \textcolor{blue}{\textbf{52.42}}\\ \midrule
 \textbf{it} 
 & 80.40 & 77.28 & 79.89 & \textcolor{blue}{\textbf{81.00}} 
 & 60.00 & 63.22 & 59.46 & \textcolor{blue}{\textbf{63.31}} 
 & 60.12 & \textcolor{blue}{\textbf{62.79}} & 55.96 & 60.36
 & 36.56 & \textcolor{blue}{\textbf{49.73}} & 31.71 & 49.62 
 & 41.14 & 54.36 & 31.71 & \textcolor{blue}{\textbf{56.18}} \\ \midrule
 \textbf{nl} 
 & 80.80 & 80.60 & 78.69 & \textcolor{blue}{\textbf{81.10}}
 & 55.96 & 60.26 & 53.27 & \textcolor{blue}{\textbf{64.12}} 
 & 57.45 & 57.84 & 56.20 & \textcolor{blue}{\textbf{59.26}} 
 & 38.18 & 49.40 & 33.76 & \textcolor{blue}{\textbf{49.83}} 
 & 39.19 & 52.66 & 32.88 & \textcolor{blue}{\textbf{55.58}} \\ \midrule
 \textbf{pt} 
 & 78.49 & \textcolor{blue}{\textbf{80.60}} & \textcolor{blue}{\textbf{80.60}} & \textcolor{blue}{\textbf{80.60}} 
 & 60.00 & 59.82 & 59.55 & \textcolor{blue}{\textbf{60.17}} 
 & 56.82 & \textcolor{blue}{\textbf{60.12}} & 55.80 & 58.47 
 & 38.40 & \textcolor{blue}{\textbf{48.22}} & 32.68 & 46.06 
 & 42.23 & \textcolor{blue}{\textbf{51.94}} & 32.76 & 51.21 \\ \midrule
 \textbf{A} & 79.93 & 79.90 & 80.25 & \textcolor{blue}{\textbf{80.96}} & 57.99 & 61.10 & 59.09 & \textcolor{blue}{\textbf{62.18}} & 58.28 & \textcolor{blue}{\textbf{59.99}} & 55.21 & 59.37 & 38.52 & \textcolor{blue}{\textbf{48.78}} & 33.09 & 48.40 & 41.42 & 52.22 & 32.21 & \textcolor{blue}{\textbf{53.72}}\\ \midrule
 \end{tabular}
 \label{results}
 \end{table*}

 \noindent\textbf{mBERT \cite{devlin2018bert}}: It was pre-trained on 102 languages in a self-supervised fashion. It makes use of two different pretext objective for pre-training: masked language modeling (MLM) and next-sentence prediction (NSP). The model through these objectives learns the inner representation of the languages in training set and later on can be used for various downstream tasks in multiple languages. 

 \noindent We use VideoMAE\footnote{\url{https://huggingface.co/docs/transformers/model_doc/videomae}}, AST\footnote{\url{https://huggingface.co/docs/transformers/model\_doc/audio-spectrogram-transformer}}, and mBERT\footnote{\url{https://huggingface.co/google-bert/bert-base-multilingual-uncased}}, openly available in Hugginface. We resample the audios to 16 Khz before passing to AST. Like previous works such as \cite{behera23_interspeech} focusing on building multilingual QA systems that uses language identifier, \textbf{MERA} uses mBERT for extraction of text representations which doesn't require a language identifier for each incoming question and eliminates the necessity of relying on the accuracy of the language identification model and the need of different models for extracting representations for different languages thus preventing computational overhead.

\vspace{-0.2cm}
\subsection{Suite of Models}\label{sec:model} 

We select the modeling networks as they are commonly used across various related tasks \cite{li2023object, khandelwal23_interspeech, tomar2023your, zaiem23b_interspeech}. 
For MERA-L, extracted representations from the foundation models are passed to the LSTM layers with hidden size of 60. We use an individual LSTM layer for each modality followed by concatenation for fusion. We add fully connected network (FCN) consisting of three layers with 200, 90, and 56 neurons followed by a classification head that represents the output answer. For MERA-C, representations from the foundation models are passed through cascaded convolution blocks with each block consisting of 32, 64, 128, 256 filters respectively with size of 3. These convolution blocks are modeled modality-wise and then concatenated before passing it to FCN. MERA-T, comprises cascaded convolution blocks for individual modality representations with similar settings that of MERA-C. These are then followed by vanilla transformer encoder \cite{vaswani2017attention} with the number of heads as 8. We use concatenation for fusion and lastly passed through FCN. FCN of MERA-C and MERA-T is of the same settings as used for MERA-L. We use ReLU as activation function in the intermediate layers. MERA-L, MERA-C, MERA-T contains 26.51, 92.84, 102.03 million trainable parameters respectively. The classification head represents the output layer with the number of neurons same as the answer labels i.e 42 (sum of the answer labels across different question types) in the case of m-MUSIC-AVQA. We do this to prevent building and training models for each individual question types as done by Li et al. \cite{li2022learning}. However, we built and train individual models for each language. We use softmax as the activation function of the classification head which outputs the probabilities for the answer labels. We exclusively conduct experiments with m-MUSIC-AVQA due to computational limitations. As AVQA comprises approximately six times more videos than MUSIC-AVQA and, training a model for each language for m-AVQA imposes significant computational overhead. However, as we are releasing both the datasets, so we encourage future studies to benchmark on it. \par

\noindent\textbf{Training Details}: We use the official training, validation, and test set given by \cite{li2022learning}. We train the models for 50 epochs on the training set with cross-entropy as the loss function and Rectified Adam as the optimizer with a learning rate of 1e-3 and batch size of 32. We also use early stopping and dropout for preventing overfitting. 
We use Tensorflow library for implementations.

\vspace{-0.3cm}
\section{Results and Analysis}
Evaluation scores of the models are shown in Table \ref{results}. The scores obtained shows the efficacy of the proposed multilingual AVQA dataset. Among the proposed models (MERA-L, MERA-C, MERA-T), MERA-C showed the best performance. MERA-C achieved good performance than MERA-T despite having lesser parameters and this can be attributed to the presence of the transformer encoder in MERA-T that may require more training data for improved performance. Among MERA-L and MERA-T, the scores were mostly similar or mixed performance where one leading in some instances and vice versa. \par

Furthermore, we experiment with ensemble of the proposed models, specifically, with weighted-ensemble of the models. Weighted-ensemble ($\hat{y}_{\text{ensemble}}$) is represented as, 
\begin{equation}
\hat{y}_{\text{ensemble}} = \alpha \hat{y}_{L} + \beta \hat{y}_{C} + \gamma \hat{y}_{T} \label{eq:ensemble}
\end{equation}
where, $\hat{y}_{L}, \hat{y}_{C}, \hat{y}_{T}$ represents the output probabilities for MERA-L, MERA-C, MERA-T respectively. With $\alpha=\beta=\gamma = 0.33$, we got the topmost performance in comparison to the individual models in most of instances, this shows the ensemble of the models is able to handle the variability in the models. However, this comes with a cost, as the inference time increases in comparison to the individual models. Inference time on the whole test set of the ensemble of models is approx 172.4 ms and it is 106.7 ms for MERA-T. MERA-T has the highest inference time in comparison to MERA-L and MERA-C. Depending on the requirement and computational power at hand, the users can choose between MERA-C and the ensemble of the models. As for comparison with previous works, there is no study exploring AVQA in multilingual settings. However, for comparison with works on the english original version of MUSIC-AVQA, we report comparable performance.

\vspace{-0.3cm}
\section{Conclusion}
We present two multilingual AVQA datasets covering eight diverse languages, generated through machine translation from existing AVQA benchmarks. We introduce \textbf{MERA}, a framework for multilingual AVQA that leverages SOTA video, audio, and textual foundation models for multilingual AVQA. Subsequently, we introduce a suite of three models (MERA-L, MERA-C, MERA-T) with different modeling architectures to benchmark the proposed multilingual AVQA datasets. Among these models, MERA-C demonstrates the best performance across various question types and exhibits robustness across the eight languages. To further enhance performance, we employ weighted-ensemble of the models, achieving the highest level of performance. Our work will serve as a benchmark study, providing a foundation for future research to evaluate their systems for multilingual AVQA. 



\bibliographystyle{IEEEtran}
\bibliography{mybib}

\end{document}